\documentclass[runningheads]{llncs}
\usepackage{arxiv}
\usepackage{amsmath}
\usepackage{booktabs} 
\usepackage{caption} 
\usepackage{subcaption} 
\usepackage{graphicx}
\usepackage{pgfplots}
\usepackage[all]{nowidow}
\usepackage[utf8]{inputenc}
\usepackage{tikz}
\usetikzlibrary{er,positioning,bayesnet}
\usepackage{multicol}
\usepackage{algpseudocode,algorithm,algorithmicx}
\usepackage{hyperref}
\usepackage{amssymb}
\usepackage[inline]{enumitem} 
\definecolor{blue}{HTML}{1F77B4}
\definecolor{orange}{HTML}{FF7F0E}
\definecolor{green}{HTML}{2CA02C}

\pgfplotsset{compat=1.14}

\author{Ashutosh ~Hathidara\thanks{Graduated from Indiana University and is no-longer associated to University for any research.} \\
	Luddy School of Informatics, Computing \& Engineering\\
	Indiana University\\
	Bloomington, IN 47405 \\
	\texttt{ashuhath@iu.edu} \\
	\And
	Lalit ~Pandey \\
	Luddy School of Informatics, Computing \& Engineering\\
	Indiana University\\
	Bloomington, IN 47405 \\
	\texttt{lpandey@iu.edu} \\
}

\begin{document}
\title{Neuro-Symbolic Sudoku Solver}
%
%


%
\maketitle 
\begin{abstract}
    Deep Neural Networks have achieved great success in some of the complex tasks that humans can do with ease. These include image recognition/classification, natural language processing, game playing etc. However, modern Neural Networks fail or perform poorly when trained on tasks that can be solved easily using backtracking and traditional algorithms. Therefore, we use the architecture of the Neuro Logic Machine (NLM) \cite{dong2019neural}  and extend its functionality to solve a 9X9 game of Sudoku. To expand the application of NLMs, we generate a random grid of cells from a dataset of solved games and assign up to 10 new empty cells. The goal of the game is then to find a target value ranging from 1 to 9 and fill in the remaining empty cells while maintaining a valid configuration. In our study, we showcase an NLM which is capable of obtaining 100\% accuracy for solving a Sudoku with empty cells ranging from 3 to 10. The purpose of this study is to demonstrate that NLMs can also be used for solving complex problems and games like Sudoku. We also analyse the behaviour of NLMs with a backtracking algorithm by comparing the convergence time using a graph plot on the same problem. With this study we show that Neural Logic Machines can be trained on the tasks that traditional Deep Learning architectures fail using Reinforcement Learning. We also aim to propose the importance of symbolic learning in explaining the systematicity in the hybrid model of NLMs.
\end{abstract}

\section{Introduction}

The groundbreaking results of the modern deep learning models have proved that they are the ideal tools to solve complex problems, however the lack of systematicity in these models have been a problem for some time. Recent research has focussed on this issue by generating hybrid models which combine Neural Networks with Symbolic Learning. In past, researchers have attempted to create hybrid models for tasks such as Language Translation \cite{bahdanau2016neural}, Curriculum Learning \cite{bengio-2009}, Learnable Recursion Logic \cite{cai2017making} and the Synthesizing complex logic based on Input-Output example pairs \cite{chen2018synthesizing}, etc. By testing one such model, called the Neural Logic Machine \cite{dong2019neural}, we emphasise on the relevance of symbolic learning in solving complex problems on which modern deep learning methods may fail. More specifically, we validate the NLM model on a different mathematical problem to realize their true potential as well as analyse their performance. Importantly NLMs can utilize the knowledge gained from generated rules to achieve a perfect generalization in several tasks. Therefore, we also aimed to test the NLMs ability to recover these lifted rules and apply them in the later stages of curriculum learning - when the complexity of the problem rises. To accomplish this, we gradually changed the number of empty cells in the grid while training.

We test the architecture of Neural Logic Machines \cite{dong2019neural} for solving a complex puzzle called Sudoku using our own set of predicates as input. In this experiment, we closely analyse the performance of this model and compare it with traditional algorithms on the same problem. To perform this experiment, we completed three main tasks. First, we trained the NLM on sudoku grids with pre-defined empty cells, the number of which increased as training progressed. This approach , where the complexity of the problem increases over training, is known as curriculum learning. Secondly, we used symbolic learning with reinforcement rewards to award the model every time a valid configuration of the empty cells was achieved. Finally, the convergence time of the NLM and Backtracking algorithm was compared using a graph plot. Towards the end of the experiment, we successful tested the model with random sudoku grids.

In the later sections, we elaborate upon the robustness of the algorithm and systematicity of the network layers.

\subsection{Key Contributions}
Our major contributions in this paper are:
\subsubsection{Extending the Applications of Neural Logic Machines:}
The NLM is trained and tested on a completely different problem set (e.g., Sudoku puzzles) to expand its scope in wide areas of applications. In \cite{dong2019neural}, linear space problems (e.g., sorting arrays) are used to test this model, whereas, this paper focuses on using 2-dimensional problem set on the same model. Instead of function approximation, the reinforcement training algorithm ‘REINFORCE’ is used to estimate the gradients using a policy-gradient mechanism and calculate the gradients in a non-differentiable trainable setting.

\subsubsection{Time Complexity and Comparison with Backtracking:}
Upon successful implementation of NLM on a 9X9 Sudoku grid, their convergence time is compared with Backtracking algorithm and demonstrated using a graphical representation. A thorough comparison of NLM with backtracking is also mentioned in the result section.

\subsubsection{Testing the Robustness of the Neural Logic Machine:}
While the NLM \cite{dong2019neural} is tested with tasks like list sorting, path finding and BlocksWorld games, we have chosen a more complex problem; solving a 9X9 sudoku puzzle with upto 10 empty cells. To sort an array, we need to compare elements with each other and swap if needed. Whereas, in Sudoku, we need to fill the gaps with appropriate numbers checking rows, columns and sub-matrices. This makes the problem more complex.

\section{Related work}

The rising demand to train Neural Networks for performing complex tasks has generated great attention among the researchers. However, their lack of systematicity and inability to generalize for a greater set of inputs has lead them to perform poorly on more systematic tasks. To address these challenges, \cite{dong2019neural} proposed the Neural Logic Machine, which can solve problems requiring systems to perform actions by following systematic sets of rules. In \cite{dong2019neural}, NLMs utilizes the relationships of objects obtained from quantifiers and logic predicates to solve BlocksWorld games, list sorting and path-finding tasks. The study done in \cite{dong2019neural} highlights the difference between conventional RNNs (Recurrent Neural Networks) with their proposed NLM, addressing the RNN’s difficulty on smaller lists, and failure to sort slightly larger lists during testing. The reason behind that is RNNs trained on smaller lists will not be able to systematically generalize it for larger lists whereas NLM can.

\paragraph{} An alternate approach to function approximators has been used with NLM called the REINFORCE algorithm \cite{NIPS1999_464d828b}, which is used for policy gradient optimization and estimates the gradient using a Monte-Carlo method. This is commonly used in deep reinforcement learning where the actions are sampled and the neural network can not perform backpropagation since sampling is non-differentiable operation. In such non-differentiable settings, we instead use gradient estimation techniques. 

In 2019, Wang et al. \cite{wang2019satnet} proposed a novel deep-learning architecture called the SATNet, which is a differentiable maximum satisfiability solver that uses CNNs. It is an approximate-differentiable solver which works on a fast coordinate descent approach for solving semidefinite programs (SDP) associated with the Maximum Satisfiability problem.

\paragraph{}While the previous studies have focused on using NLMs on different problem sets \cite{dong2019neural} or solving Sudoku puzzles with fully Deep Learning approaches \cite{wang2019satnet}, our experiment emphasizes the combination of Symbolic Learning with Deep Learning, as well as a hybrid architecture to solve a new sets of complex problems. Lastly, this experiment also focuses on realizing the true potential of NLMs in different areas of applications.

\section{Proposed Methodology}
\begin{figure}
    \centering
    \includegraphics[width=15cm, height=6cm]{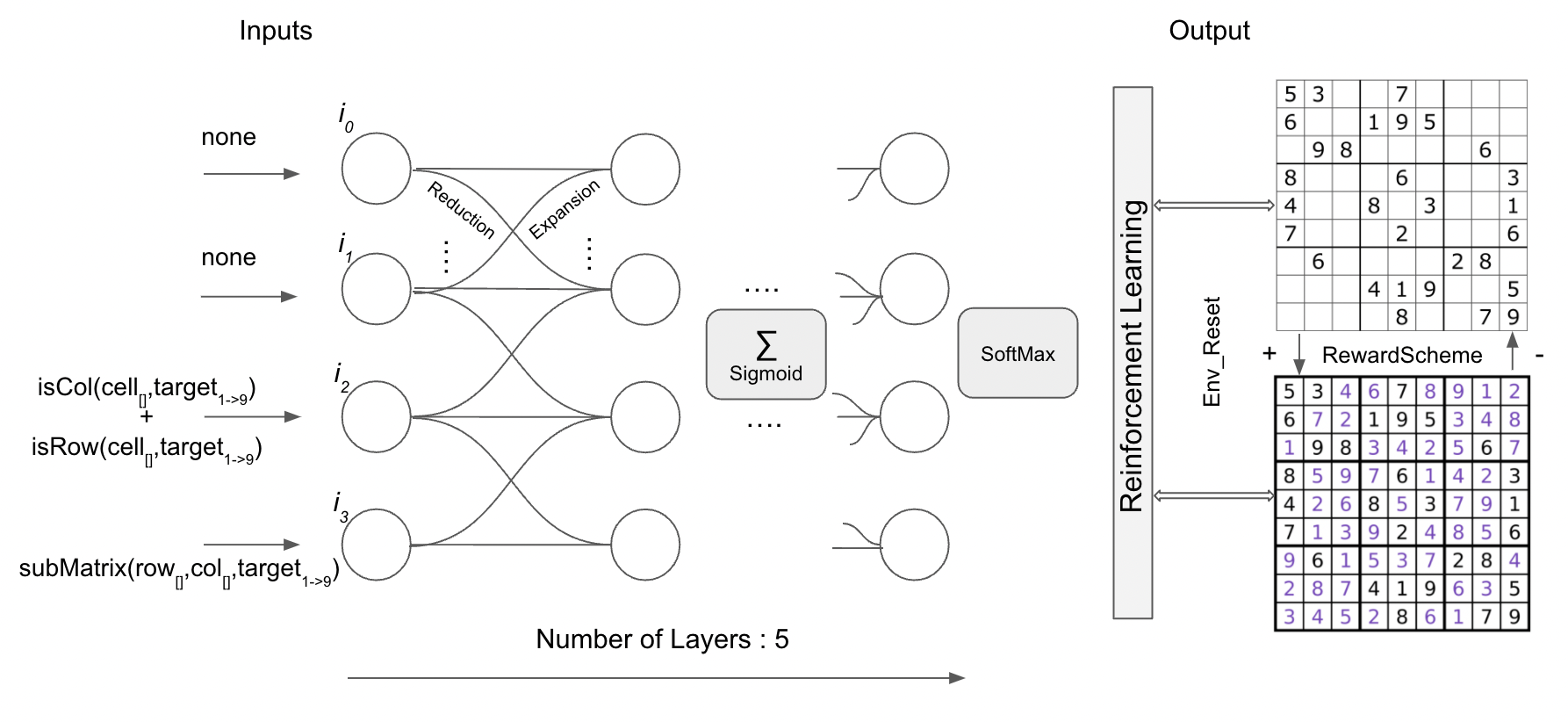}
    \caption{Model Architecture of Neural Logic Machine}
    \label{fig:my_label}
\end{figure}

\paragraph{}Figure 1 illustrates our proposed model architecture for the Neuro-Symbolic Sudoku Solver - a Hybrid architecture of Deep Reinforcement Learning techniques and Symbolic Learning. The first half of the model constitutes the learning phase, where the Sigmoid function acts as the activation function between hidden layers, and the SoftMax function activates the output layer. The input layer consists of 4 neurons each accepting a certain type of parameter. The flexibility to allocate a certain type of input to each neuron, leads to greater systematicity in the model. For instance, out of the four neurons in the input layer, the first neuron, $i_0$, accepts only a nullary predicate; $i_1$ accepts a unary predicate; $i_2$ accepts a binary predicate, and $i_3$ receives a ternary predicate as an input.

The problem of solving Sudoku puzzles requires checking for each row, column, and submatrix to maintain a valid configuration of the Sudoku grid. In order to check this constraint, the coordinates of the rows and columns along with the target values are passed to the input layer. Therefore, neurons $i_2$ and $i_3$ receive input as binary and ternary predicates, while neurons $i_0$ and $i_1$ receive Null inputs. In contrast to this experiment, \cite{dong2019neural} shows how their problems require predicates for a different set of neurons in the NLM. For example, \cite{dong2019neural} uses $i_1$ and $i_2$ since unary and binary predicates are required to solve the array sorting problem. Once the set of predicates is received by the input layer, the input for the following layers are reduced or expanded based on the arity of the previous layer.

The output from the SoftMax layer is fetched by the Reinforcement Learning
(RL) module, which constitutes Phase 2 of the architecture (see figure
2). The RL module takes care of three main functionalities: Allocating a +1 positive reward for a fully solved grid, a negative reward of -0.01 for every move and performing environmental resets after checking all target values. The RL Module will trigger an environmental reset if none of the target values from 1-9 can fill an empty cell while maintaining a valid configuration of the Sudoku grid. During this reset, all filled values are emptied, the sudoku grid is reinitialized, and a new iteration begins.

\begin{figure}
    \centering
    \includegraphics[width=12cm, height=8cm]{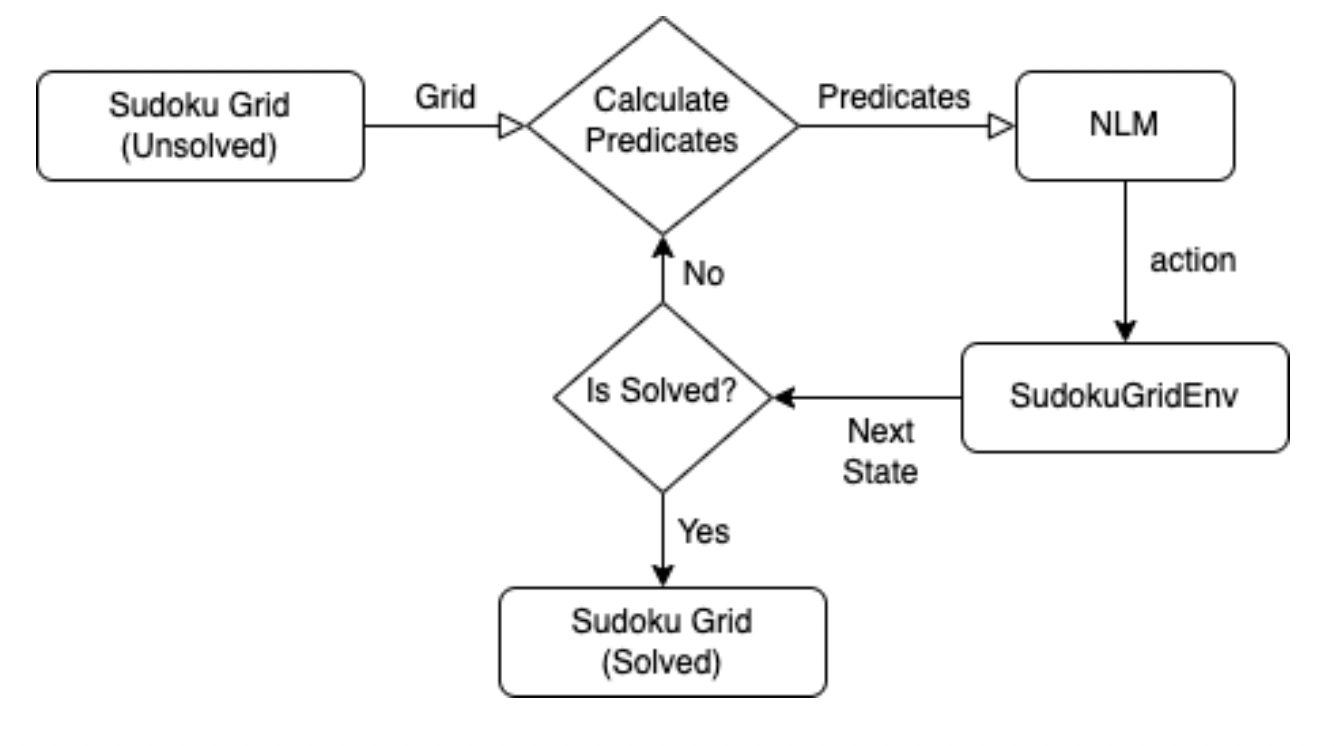}
    \caption{High Level System Architecture}
    \label{fig:my_label}
\end{figure}

\paragraph{}Given an unsolved Sudoku grid, we have implemented a multistage architecture to solve the grid step-by-step. Figure 2 illustrates the high-level architecture of the Neuro-Symbolic Sudoku solver, which aligns with problems experimented in the original NLM paper.

The first step in this implementation diagram consists of calculating the boolean predicates. There are three important rules to solve a sudoku grid: The solver needs to put a number in an empty cell such that the resulting grid configuration remains valid. Here, valid configurations refers to the states where each number in every row, column, and 3x3 submatrix is distinct. With the help of these conceptual rules, lifted Boolean predicates may be created.

Lifted rules are generalized rules which apply to a task to crack its solvability. These rules are applicable to any example in that task domain, irrespective of its configuration or complexity. These rules can be seen as the simplest fundamental rules of solving a system. We define the predicate isRow(r, x), which computes whether number x exists anywhere in row r. Similarly, predicate isColumn(c, x) computes whether number x exists anywhere in column c and predicate isSubMat(r, c, x) computes whether number x exists in the 3x3 submatrix containing the cell (r, c). Based on above definition of the predicates, isRow(r, x) and isColumn(c, x) are binary predicates and both result in [9, 9] shaped tensors whereas isSubMat(r, c, x) is a ternary predicate that results in a [9, 9, 9] shaped tensor.

It is also worth mentioning that in this study we do not input the unsolved grid into the input layer of the neural network. Instead, we compute the predicates as described above and pass those predicates as a set of input. Since there are two binary predicates, we concatenate the values of isRow(r, x) and isColumn(c, x) and call the resultant tensor as a stacked binary predicate. Now, these predicates can be feeded into the input layer of the neural network.

The last layer of the NLM logic machine computes the SoftMax value and provides the empty cell position, (r, c), as well as the target value to place in the empty cell. Here comes the role of the reinforcement module. The Reinforcement module checks if placing x into (r, c) makes a valid sudoku grid or not. Based on the previous assertion, it generates the next state and computes the positive reward (if the next state is not a valid sudoku grid) or negative reward (if the next state is not a valid sudoku grid). The reinforcement algorithm also checks if the next state generated is a solved Sudoku grid. In this case, we break from the iteration and print the output, otherwise, we repeat the same steps.

However, the above strategy may take indefinite number of steps to find a solution. To prevent this, we have defined an upper bound on the number of steps the algorithm may take. The proposed algorithm yields a success rate of 1 if it solves the grid, and a 0 otherwise.

\subsection{Training Details}

\begin{table}[H]
\centering
\begin{tabular}{|l|l|l|l|l|l|l|l|l|l}
\hline
                             \textbf{Problem} &
                             \textbf{Epochs} &
                             \textbf{Batch Size} &
                             \textbf{Loss} &
                             \textbf{Optimizer} &
                             \textbf{LR} &
                             \textbf{RL Rewards} &
                             $\boldsymbol{\gamma}$
                             \\ \hline
    \textbf{Sudoku Puzzle}          & 50            & 4              & Softmax-CE      & Adam   &0.005      & +1,-0.01    & 0.99            \\ \hline
    
            \end{tabular}
            \vspace{4mm}
        \caption{Training Details of Neuro-Symbolic Sudoku Solver}
\end{table}

The hyper-parameters and the training details of the NLM for solving a 9x9 sudoku puzzle are shown in table 1.

\section{Result and Analysis}

\begin{table}[h]
  \begin{center}
    \label{tab:table1}
    \begin{tabular}{c|c|c} 
      \textbf{No. of Empty Cells} & \textbf{Max. Steps} & \textbf{Success Rate}\\
      \hline
      3 & 81 & 0.94\\
      3 & 150 & 1.00\\
      3 & 400 & 1.00\\
      3 & 729 & 1.00\\
      5 & 81 & 0.80\\
      5 & 150 & 0.96\\
      5 & 400 & 0.99\\
      5 & 729 & 1.00\\
      8 & 80 & 0.68\\
      8 & 150 & 0.92\\
      8 & 400 & 0.98\\
      8 & 729 & 1.00\\
    \end{tabular}
  \end{center}
  \caption{Comparison of different training parameters}
\end{table}

Our experiment involves multiple settings (number of empty cells, dimensions and optimal steps) of the grid, which are often modified during the testing phase to understand the performance of the model and obtain a pattern from the results. To begin the experiment, the number of empty cells and the maximum steps in the sudoku grid are limited to 3 and 81 respectively. These are then gradually incremented as the model trains. As these parameters change over time, the complexity to solve the problem also increases. However, our result suggests that NLMs can perfectly confront this complexity and yields 100\% accuracy in most of the cases.

To give a better understanding on how the model performs with different parameters, we have demonstrated the success rate with respect to each parameter that the model was trained on. Table 1 shows the comparison and performance under each setting that is tested on our modified version of the NLM. The model when tested with the minimum number of empty cells (nr empty) 3 and max steps set to 81, gives a success rate of 0.94. However, when the maximum number of steps (max steps) are increased from 81 to 150, the model receives a perfect score of 1.

A similar case is also observed when the model receives 5 empty cells with
81 and 729 max steps. With fewer optimal steps, the model yields a comparatively lower score as compared to 3 empty cells. However, when the model does not have a restriction on the maximum number of steps (set to a max of 729), it performs well and gets 100\% accuracy. From this we conclude that success rate is directly related to the maximum number of steps allowed by the model. Another inference obtained from Table 2 is that when number of empty cells increases keeping Max. steps constant, success rate drops. This shows the inverse relation between success rate and empty cells. Therefore, the success rate of NLM model is strongly determined by the relation of:

\begin{equation}
    \textit{Success Rate} \boldsymbol{\propto} \textit{max steps} * \textit{$\frac{1}{empty cells}$}
\end{equation}

\begin{figure}
    \centering
    \includegraphics[width=12cm, height=8cm]{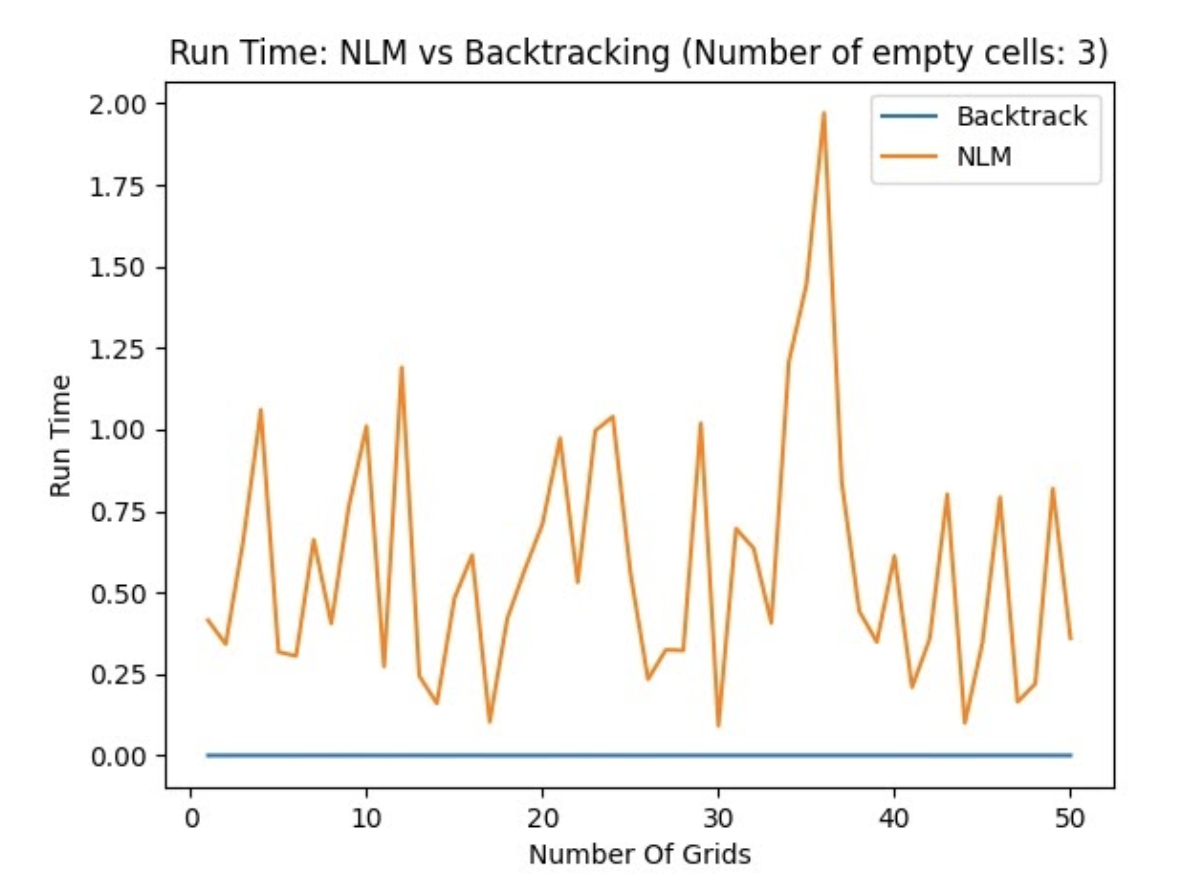}
    \caption{Comparison of NLM and backtracking convergence time}
    \label{fig:my_label}
\end{figure}

\begin{figure}
    \centering
    \includegraphics[width=15cm, height=8cm]{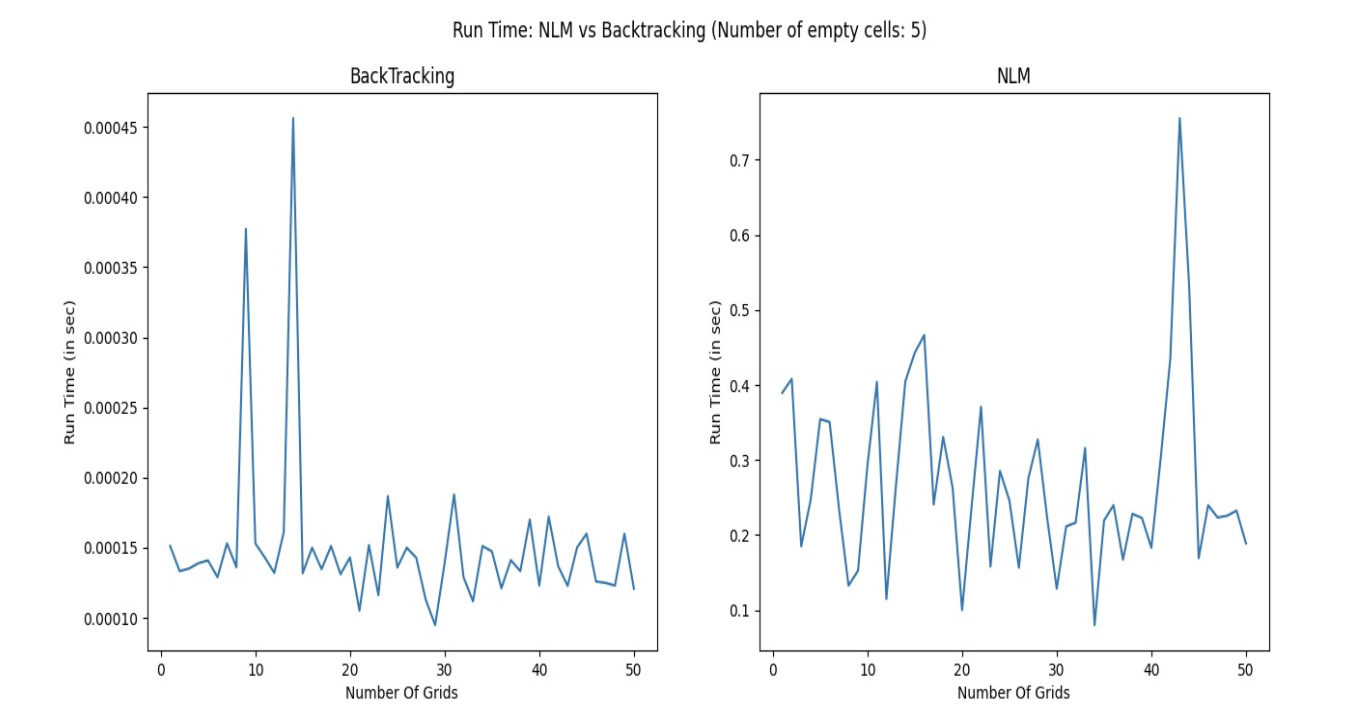}
    \caption{Separate convergence time for Backtracking and NLM for same problem}
    \label{fig:my_label}
\end{figure}

In addition to fine-tuning the model, we have also drawn a time complexity analysis of the NLM with traditional backtracking algorithms for solving Sudoku puzzles. Both the NLM and backtracking algorithms are provided with the same set of grids and their time to solve the complete grid is highlighted in Figure 3. The motivation behind this is to showcase the difference in the principle working of both the algorithms and analyze their convergence time (with limited training in the case of NLMs).
\paragraph{}
The backtracking algorithm takes a constant average time of nearly 0.00045 seconds to solve the same set of 9X9 grids, on the other hand, the NLM takes a considerably higher amount of time to converge. (It is also worth mentioning that Figure 3 demonstrates the time taken by the NLM with 729 maximum number of steps). The reason that backtracking converges faster is due to the fact that it solves the grid in an optimal number of steps (i.e., $MaxStepsBacktracking = NumberOfEmptyCells$). In Figure 3, the peak time in the case of the NLM denotes the instances in which the environment was reset due the formation of an invalid configuration of the sudoku grid. During this instance, the model first receives a negative reward through the reinforcement module and then resets the environment once there are no possible target values to test. In this case, the NLM again tries to fill the empty cells but with a different set of target values from the beginning. However, even with 10 empty cells, our modified version of the NLM always takes less than 2.0 seconds to converge.

\section{Conclusion and Future Work}
The focus of this study is to tackle one of the drawbacks of the traditional Neural Networks i.e., ‘systematicity’. Where Neural Networks perform poorly, NLMs can solve  some the same task with 100\% accuracy. NLMs have been trained and tested by \cite{dong2019neural} on various tasks which Deep Learning models have failed to solve or converge. In our paper, we added to the existing  applications of their architecture and solved a more complex problem to test the robustness of Neuro Logic Machines.

While the Neuro Logic Machines failed to converge for Sudoku puzzles faster than the backtracking algorithm, it is evident from this study that a Neuro Logic Machine can be trained to solve tasks where conventional Deep Learning models may fail. Lastly, because the NLM receives a random combination of grids and number empty cells from, we are confident that the high success rate of NLMs is not due to the model’s over-fitting. Thus, with this experiment, we have been able to strengthen the argument \cite{dong2019neural} that NLM can solve tasks with 100\% accuracy without relying on over-fitting. In Section 4, we also deduce that that the success rate is directly associated with the number of empty cells and the maximum number of steps that model is allowed to take.

To conclude, Neuro Logic Machines can solve complex problems using a hybrid approach of Reinforcement and Symbolic Learning. In future work, we intend to cover the fine-tuning and convergence rate of the algorithm. We propose that the applications of NLMs can be extended further with even more games (e.g., Ken Ken puzzles) and mathematical problems (such as search tasks). We also anticipate that the architecture will cover problems where NLM have not yielded a success rate of 100\%.

\section{Acknowledgment}
We thank Dr. Leake, professor at Indiana University Bloomington, for being our instructor and guiding us through this experiment and thereby supporting our work.

\subsection{Conflict of Interest}
The authors declare that they have no conflict of interest.

\subsection{Funding}
The authors received no financial support for the research, authorship, and/or publication of this article.

%
%
%
\bibliographystyle{plain} 
\bibliography{biblio} 


%
\end{document}